\documentclass{article}
\usepackage{ijcai19}

\usepackage{times}
\usepackage{soul}
\usepackage{url}
\usepackage[utf8]{inputenc}
\usepackage[small]{caption}
\usepackage{graphicx}
\usepackage{amsmath}
\usepackage{amsfonts}
\usepackage{booktabs}
\usepackage{algorithm}
\usepackage{algorithmic}
\newtheorem{theorem}{Theorem}
\newtheorem{lemma}{Lemma}

\newtheorem{assumption}{Assumption}
\newtheorem{remark}{Remark}
\newtheorem{proof}{Proof}
\usepackage{subcaption}

\title{Quadruply Stochastic Gradients for Large Scale Nonlinear \\ Semi-Supervised AUC Optimization}

\author{
Wanli Shi$^{1}$
\and
Bin Gu$^{1,2}$\and
Xiang Li$^{3}$\and
 Xiang Geng$^{1}$\And
Heng Huang$^{2,4}$
\footnote{Contact Author}
\affiliations
$^{1}$School of Computer \& Software, Nanjing University of Information Science \& Technology, P.R.China\\
$^{2}$JD Finance America Corporation\\
$^{3}$Computer Science Department, University of Western Ontario, Canada\\
$^{4}$Department of Electrical \& Computer Engineering, University of Pittsburgh, USA
\emails
wanlishi@nuist.edu.cn,
jsgubin@gmail.com,
lxiang2@uwo.ca,
gengxiang@nuist.edu.cn,
heng.huang@pitt.edu
}

\begin{document}

\maketitle

\begin{abstract}
	Semi-supervised learning is pervasive in real-world applications, where only a few labeled data are available and large amounts of instances remain unlabeled. Since AUC is an important model evaluation metric in classification, directly optimizing AUC in semi-supervised learning scenario has drawn much attention in the machine learning community. Recently, it has been shown that one could find an unbiased solution for the semi-supervised AUC maximization problem without knowing the class prior distribution. However, this method is hardly scalable for nonlinear classification problems with kernels. To address this problem, in this paper, we propose a novel scalable quadruply stochastic gradient algorithm (QSG-S2AUC) for nonlinear semi-supervised AUC optimization. In each iteration of the stochastic optimization process, our method randomly samples a positive instance, a negative instance, an unlabeled instance and their random features to compute the gradient and then update the model by using this quadruply stochastic gradient to approach the optimal solution. More importantly, we prove that QSG-S2AUC can converge to the optimal solution in $O(1/t)$, where $t$ is the iteration number. Extensive experimental results on  a variety of benchmark datasets show that QSG-S2AUC is far more efficient than the existing state-of-the-art algorithms for semi-supervised AUC maximization, while retaining the similar generalization performance.
\end{abstract}

\section{Introduction}

Semi-supervised learning addresses the problems where the available data is composed of a small size of labeled samples and a huge size of unlabeled samples. It is of immense practical interest in a wide range of applications, such as image retrieval ~\cite{wang2010semi}, natural language processing ~\cite{liang2005semi} and speech analysis ~\cite{sholokhov2018semi}. Since semi-supervised learning requires less human effort and can achieve a better generalization performance, it has attracted a great deal of attention in the machine learning communities, \textit{i.e.}, ~\cite{gu2018new,sakai2017semi,sakai2018semi,geng2019scalable,Yu2019tackle}.


 The area under the ROC curve (AUC) ~\cite{hanley1982meaning} measures the probability of a randomly drawn positive instance being ranked higher than a randomly drawn negative instance. Thus, AUC is a  more effective performance measure than the accuracy in data imbalance binary classification problem. Many studies ~\cite{gao2013one,gao2015consistency} have also pointed out that optimizing AUC can achieve a better generalization performance than directly optimizing accuracy. Due to the superiority of AUC as mentioned above, a large amount of attention has been attracted to introduce AUC to semi-supervised learning.

Recently, several algorithms have been proposed to address the semi-supervised AUC optimization problem. For instance, to train a classifier, SSRankBoost ~\cite{amini2008boosting} and OptAG ~\cite{fujino2016semi} exploited the assumption that two samples share the same label if their distance in a metric space is small. However, this restrictive assumption may not always hold in real-world applications, and could lead to biased solutions. Sakai et al., \shortcite{sakai2018semi} pointed out that both unlabeled instances and labeled instances follow the same joint probability distribution and the restrictive assumption is not necessary. However, their method PNU-AUC requires to estimate the class prior which is difficult to be obtained when labeled instances are extremely small. Recently, Xie and Li, \shortcite{xie2018semi} proposed that neither the class priors nor any other distributional assumption about the unlabeled data are necessary to find the unbiased solution. We summarize these algorithms in Table \ref{tab:SS_AUC_Optimization}.

\begin{table*}
	\centering
	\setlength{\tabcolsep}{2mm}

	\begin{tabular}{lllcc}
		\toprule
		\textbf{Algorithm}  & \textbf{Reference}&\textbf{Function model}  & \textbf{Computational complexity} & \textbf{Space complexity} \\
		\hline
		SSRankBoost&~\cite{amini2008boosting}&Nonlinear model &--- &$O(n^2)$ \\
		OptAG & ~\cite{fujino2016semi} & Linear model &--- &$O(n^2)$	\\
		PNU-AUC & ~\cite{sakai2018semi}& Nonlinear model& $O(n^3)$ & $O(n^2)$\\
		SAMULT & ~\cite{xie2018semi}  &Nonlinear model&$O(n^3)$&$O(n^2)$ \\
		\hline
		QSG-S2AUC &  Ours&Nonlinear model &$O(Dt^2)$&$O(t)$\\
		\bottomrule
	\end{tabular}

	\caption{Several representative semi-supervised AUC optimization algorithms. ($D$ denotes the number of random features, $n$ denotes the number of training samples and $t$ denotes number of iterations.)   }
	\label{tab:SS_AUC_Optimization}
\end{table*}

Nonlinear data structures widely exist in many real-world problems, and kernel method is a typical way to solve such problems \cite{huang2019faster,gu2014incremental}. However, this approach can hardly scale to large datasets. Specifically, the kernel matrix needs $O(n^2d)$ operations to be calculated and $O(n^2)$ to be stored, where $n$ denotes the number of instances and $d$ denotes the dimensionality of the data ~\cite{gu2018asynchronousg}. However, the bottlenecks of the computational complexities become more severe for semi supervised learning because the sample size $n$ is
always very large in the semi-supervised
scenario. Even worse, PNU-AUC and SAMULT ~\cite{xie2018semi} need $O(n^3)$ operations to compute the matrix inverse. Thus, scaling up non-linear semi-supervised AUC maximization is a challenging problem.

To scale up kernel-based algorithms, a large amount of methods has been proposed, \textit{i.e.}, asynchronous parallel algorithms ~\cite{gu2018accelerated,gu2018asynchronousg,gu2016asynchronousb}, kernel approximation ~\cite{rahimi2008random,smola2000sparse}. To our knowledge, doubly stochastic gradient (DSG) ~\cite{dai2014scalable} is the most effective method to scale up kernel-based algorithms. Specifically, DSG samples a random  instance and the random features to compute the doubly stochastic gradient which is used to update the model. However, different from the standard DSG, semi-supervised learning has three sources of data, \textit{i.e.},  positive instances, negative instances and unlabeled datasets. In addition, optimizing AUC is a pairwise learning problem which is more complicated than the pointwise learning problem considered in the standard DSG algorithm. Therefore, the existing  algorithms and theoretical analysis for DSG cannot be directly applied to non-linear semi-supervised AUC maximization.

To address this challenging problem, we introduce multiple sources of randomness. Specifically, we randomly sample a positive, a negative and an unlabeled instance in each iteration to compose a triplet of data points. Then we use the random features w.r.t these data triplets to compute the stochastic gradient. Since the stochastic gradient would then contain four sources of randomness, we denote our algorithm as quadruply stochastic semi-supervised AUC maximization (QSG-S2AUC). Theoretically, we prove that QSG-S2AUC can converge to the optimal solution at the rate of $O(1/t)$, where $t$ is the number of gradient iterations. Extensive experimental results on  a variety of benchmark datasets show that QSG-S2AUC is far more efficient than the existing state-of-the-art algorithms for semi-supervised AUC maximization, while retaining the similar generalization performance.

\noindent \textbf{Contributions.} The  main contributions of this paper are summarized as follows.
\begin{enumerate}

\item We propose an efficient nonlinear semi-supervised AUC optimization algorithm based on the DSG framework. Since semi-supervised learning contains three sources of data, we employ triplets of data points in each iteration and extend the standard DSG framework.

\item We prove that QSG-S2AUC has the convergence rate of $O(1/t)$ which is same to  the one of standard SGD even though our QSG-S2AUC has four sources of randomness.

\end{enumerate}

\section {Related Works}
In this section, we give a brief review of kernel approximation and  large scale AUC maximization methods respectively.

%
\subsection{Kernel Approximation}
Kernel approximation has attracted great amounts of attention to scale up kernel-based learning algorithms. The data-dependent methods, such as greedy basis selection techniques ~\cite{smola2000sparse}, incomplete Cholesky decomposition ~\cite{fine2001efficient}, Nystr{\"o}m method ~\cite{drineas2005nystrom}, utilize the given training set to compute a low-rank approximation of the kernel matrix. However, they need a large amount of training instances to achieve a better generalization. To handle this challenge, random Fourier feature (RFF) ~\cite{rahimi2008random} directly approximates the kernel function unbiasedly with some basis functions. However, large amounts of memory are required since the number random features $D$ need to be larger than the original features to achieve low approximation error. To further improve RFF, Dai \textit{et al.}, \shortcite{dai2014scalable} proposed DSG algorithm. It uses \textit{pseudo-random number generators} to calculate the random features on-the-fly, which highly reduces the memory requirement. These methods have been widely applyed to scale up kernel-based learning algorithms, such as ~\cite{li2017triply,gu2018asynchronousk}. 



\subsection{Large Scale AUC Optimization}
Recently, several efforts have been devoted to scale up the AUC optimization. For example, Ying \textit{et al}., \shortcite{ying2016stochastic} formulated the AUC optimization as a convex-concave saddle point problem and proposed a stochastic online method (SOLAM) which has the time and space complexities of one datum. FSAUC ~\cite{liu2018fast} developed a multi-stage scheme for running primal-dual stochastic gradient method with adaptively changing parameters. FSAUC has the convergence rate of $O(1/n)$, where $n$ is the number of random samples. However, both SOLAM and FSAUC focus on scaling up the linear AUC optimization and are incapable of maximizing  AUC in the nonlinear setting. Recently, FOAM and NOAM ~\cite{ding2017large} used RFF and Nystr{\"o}m method, respectively, to scale up the kernel based AUC optimization problem. However, as  mentioned above, both  methods require large amounts of memory to achieve a better generalization performance and not trivial to scale up the nonlinear semi-supervised AUC optimization problems based.


\section {Preliminaries}
\subsection{Supervised AUC Optimization}

In supervised learning, 
let $x \in \mathbb{R}^d$ be a $d$-dimensional pattern and $y \in \{+1,-1\}$ be a class label. Let $p(x,y)$ be the underlying joint density of $(x,y)$. The $\mathrm{AUC}$ optimization is to train a classifier $f$ that maximizes the following function.
\begin{align}
\mathrm{AUC} &=& 1-\mathbb{E}_{x^p \sim p^{+}(x)}\left[\mathbb{E}_{x^n \sim p^{-}(x)}[l_{01}(f(x^p),f(x^n))]\right] ,\nonumber
\end{align}
where $p^{+}(x) = p(x|y=+1)$, $p^{-}(x)=p(x|y=-1)$ and $l_{01}(u,v)= (1-\mathrm{sign}(u-v))/2$. Obviously, maximizing AUC is equivalent to minimizing the following PN AUC risk.
\begin{align}\label{PN_risk}
R_{\mathrm{PN}} &= \mathbb{E}_{x^p \sim p^{+}(x)}\left[\mathbb{E}_{x^n \sim p^{-}(x)}[l_{01}(f(x^p),f(x^n))]\right].
\end{align}

Given the positive and negative datasets as $D_p = \{x_i\}_{i=1}^p \sim p^{+}(x)$ and $D_n = \{x_j\}_{j=1}^n \sim p^{-}(x)$ respectively.
Thus, the PN AUC risk can be rewritten as follows.
\begin{align}\label{empirical_PN_risk}
R_{\mathrm{PN}} &= \mathbb{E}_{x^p \in D_p}\left [\mathbb{E}_{x^n \in D_n}[l(f(x^p),f(x^n))]\right].
\end{align}
where $\mathbb{E}_{x^p \in D_p}$ and $\mathbb{E}_{x^n \in D_n}$ denote the means of $D_p$ and $D_n$, respectively.

\subsection{Semi-Supervised AUC Optimization}
 Since large amounts of instances remain unlabeled in semi-supervised learning, we assume that the labeled dataset is limited while the unlabeled data can be infinite and has the underlying distribution density of $p(x)$, where $p(x) =\pi p^{+}(x)+(1-\pi)p^{-}(x)$ and $\pi$ denotes the positive class prior.  
Recently, Xie and Li, \shortcite{xie2018semi} have shown that  it is unnecessary to estimate distributional assumptions or class prior to achieve an unbiased solution for  semi-supervised AUC optimization. Specifically, PU AUC risk $R_{\mathrm{PU}}$ and NU AUC risk $R_{\mathrm{NU}}$ are equivalent to the supervised PN AUC risk $R_{\mathrm{PN}}$ risk with a linear transformation, where PU AUC risk $R_{\mathrm{PU}}$ is estimated by positive and unlabeled data treated as negative data, and NU AUC risk $R_{\mathrm{NU}}$ is estimated by negative and unlabeled data treated as positive data. 
We define $R_{\mathrm{PU}}$ and $R_{\mathrm{NU}}$ as follows,
\begin{eqnarray}
R_{\mathrm{PU}} &=& \mathbb{E}_{x^p \in D_p}\left[\mathbb{E}_{x^u \sim p(x)}[l(f(x^p), f(x^u))]\right],
\\ 
R_{\mathrm{NU}} &=& \mathbb{E}_{x^u \sim p(x)}\left[\mathbb{E}_{x^n \in D_n}[l(f(x^u), f(x^n))]\right],
\end{eqnarray}
where $\mathbb{E}_{x^u \sim p(x)}$ denotes the expectation over the density $p(x)$. PU AUC risk can be written as follows.
\begin{align}
R_{\mathrm{PU}} = &\mathbb{E}_{x^p \in D_p}[\mathbb{E}_{x^u \sim p(x)}[l(f(x^p), f(x^u))]] \nonumber\\
= &\mathbb{E}_{x^p \in D_p}[\pi\mathbb{E}_{x'^p\sim p^{+}(x)}[l(f(x^p,x'^p))] \nonumber
\\ &+ (1-\pi) \mathbb{E}_{x'^n \sim p^{-}(x)}[l(f(x^p,x'^n))]] \nonumber\\
= &\dfrac{1}{2}\pi + (1-\pi)R_{\mathrm{PN}},
\end{align}
where $x'^p$ and $x'^n$ denotes the positive and negative instances in unlabeled dataset.
Similarly, NU AUC risk $R_{\mathrm{NU}}$ can be rewritten as
\begin{align}
R_{\mathrm{NU}}=\dfrac{1}{2}(1-\pi) + \pi R_{\mathrm{PN}}.
\end{align}
Then PN AUC risk $R_{\mathrm{PN}}$ can be formulated as follows.
\begin{align}
R_{\mathrm{PU}}+ R_{\mathrm{NU}}-\dfrac{1}{2} =  R_{\mathrm{PN}}.
\end{align}
Thus, the semi-supervised AUC optimization can be formulated as follows.
\begin{align}\label{PNU_risk}
R_{\mathrm{PNU}}=(1-\gamma)\left (R_{\mathrm{PU}}+ R_{\mathrm{NU}}-\dfrac{1}{2} \right ) + \gamma R_{\mathrm{PN}}.
\end{align}
where $\gamma \in [0,1]$ is the trade-off parameter. To reduce the risk of overfitting, we add a $l_2$-regularizer into (\ref{PNU_risk}) and have the following objective for semi-supervised AUC optimization.
\begin{align}\label{objective_function}
\mathcal{L}= R_{\mathrm{PNU}}(f)+\dfrac{\lambda}{2}\|  f \|_\mathcal{H}^2,
\end{align}
where $\lambda$ is the regularized parameter and $\| \cdot \|_{\mathcal{H}}$ denotes the norm in a reproducing kernel Hilbert space (RKHS) $\mathcal{H}$. 

\subsection{Random Fourier Feature}
In this section, we give a brief review of RFF. Assume that we have a \textit{continuous}, \textit{real-valued}, \textit{symmetric} and \textit{shift-invariant} kernel function $k(x,x')$. According to Bochner Theorem ~\cite{rudin2017fourier}, this kernel function is positive definite  and has a nonnegative Fourier transform function as $k(x,x') = \int_{\mathbb{R}^d} p(\omega)e^{j\omega^T(x-x')}d\omega$, where $p(w)$ is a density function associated with $k(x,x')$. The integrand $e^{j\omega^T(x-x')}$ can be replaced with $\cos\omega^T(x-x')$ ~\cite{rahimi2008random}. Then we can obtain a real-valued feature map $\phi_{\omega_i}(x) = [\cos (\omega_i^Tx), \sin (\omega_i^Tx) ]^T$, where $\omega_i$ is randomly sampled according to the density function $p(\omega)$. We can obtain the feature map for $m$ random features of a real-valued kernel as follows.
\begin{eqnarray}
\phi_{\omega}(x) = \sqrt{1/D}[\cos (\omega_1^Tx),\cdots,\cos (\omega_m^Tx),\nonumber\\ \sin(\omega_1^Tx),\cdots,\sin(\omega_m^Tx) ]^T .
\end{eqnarray}
Obviously, $\phi^T_{\omega}(x)\phi_{\omega}(x')$ is an unbiased estimate of $k(x-x')$.

\section{Quadruply Stochastic Semi-Supervised AUC Maximization}
\subsection{Quadruply Stochastic Gradients}
Based on the definition of the function $f \in \mathcal{H}$, we easily obtain $\nabla f(x) = \nabla\langle f,k(x,\cdot)\rangle$, and $\nabla \| f \|_\mathcal{H}^2=\nabla\langle f,f \rangle_\mathcal{H} = 2f$. Thus, the gradient of the objective (\ref{objective_function}) can be written as:
\begin{align}\label{functional_gradient}
\nabla \mathcal{L} = & \lambda f + \gamma \mathbb{E}_{x^p \in D_p}[\mathbb{E}_{x^n \in D_n}[l_1'k(x^p,\cdot)+l_2'k(x^n,\cdot)]]\nonumber\\
&+(1-\gamma) (\mathbb{E}_{x^p \in D_p}[\mathbb{E}_{x^u \sim p(x)}[l_3'k(x^p,\cdot) +l_4'k(x^u,\cdot)]]\nonumber\\
&+ \mathbb{E}_{x^u \sim p(x)}[\mathbb{E}_{x^n \in D_n}[l_5'k(x^u,\cdot)
+l_6'k(x^n,\cdot)]]),
\end{align}
where $l_1'k(x^p,\cdot)$ denotes the derivative of $l(f(x^p),f(x^n))$ w.r.t. $f(x^p)$, $l_2'k(x^n,\cdot)$ denotes the derivative of $l(f(x^p),f(x^n))$ w.r.t. $f(x^n)$, $l_3'k(x^p,\cdot)$ denotes the derivative of $l(f(x^p),f(x^u))$ w.r.t. $f(x^p)$, $l_4'k(x^u,\cdot)$ denotes the derivative of $l(f(x^p),f(x^u))$ w.r.t. $f(x^u)$, $l_5'k(x^u,\cdot)$ denotes the derivative of $l(f(x^u),f(x^n))$ w.r.t. $f(x^u)$ and $l_6'k(x^n,\cdot)$ denotes the derivative of $l(f(x^u),f(x^n))$ w.r.t. $f(x^n)$.

In order to update the classifier $f$ in a stochastic manner,
we randomly sample a positive data point $x^p$ and a negative data point $x^n$ from $D_p$ and $D_n$, respectively. In addition, we randomly sample an unlabeled data point $x^u$ according to the unlabeled data distribution density $p(x)$. In each iteration, we use a triplet of these data points to compute the stochastic functional gradient of (\ref{PNU_risk}) as follows.
\begin{align}\label{stochastic_functional_gradient}
\xi(\cdot) =& \gamma (l_1'k(x^p,\cdot) +l_2'k(x^n,\cdot))
+(1-\gamma) (l_3'k(x^p,\cdot)\nonumber\\ &+l_4'k(x^u,\cdot)+ l_5'k(x^u,\cdot)+l_6'k(x^n,\cdot)).
\end{align}

We can apply the random Fourier feature method to further approximate the stochastic functional gradient $\xi(\cdot)$ as follows.
\begin{align}\label{quadruply_stochastic_functional_gradient}
\zeta(\cdot) = & \gamma( l_1'\phi_{\omega}(x^p)\phi_{\omega}(\cdot) +l_2'\phi_{\omega}(x^n)\phi_{\omega}(\cdot))\nonumber\\
&+(1-\gamma) (l_3'\phi_{\omega}(x^p)\phi_{\omega}(\cdot) +l_4'\phi_{\omega}(x^u)\phi_{\omega}(\cdot)\nonumber\\
&+ l_5'\phi_{\omega}(x^u)\phi_{\omega}(\cdot)+l_6'\phi_{\omega}(x^n)\phi_{\omega}(\cdot)).
\end{align}
Obviously, we have that $\xi(\cdot) = \mathbb{E}_{\omega}[\zeta(\cdot)]$. Thus, we can  achieve the unbiased estimate of the gradient (\ref{functional_gradient}) by using either  $\xi(\cdot)$ or  $\zeta(\cdot)$ as follows,
\begin{align}
\nabla \mathcal{L} = \mathbb{E}_{x^p,x^n,x^u}[\xi(\cdot)]+\lambda f ,\nonumber
\nabla \mathcal{L} = \mathbb{E}_{x^p,x^n,x^u}[\mathbb{E}_{\omega}[\zeta(\cdot)]] + \lambda f. \nonumber
\end{align}
Because four randomness (\textit{i.e.} $x^p$, $x^n$, $x^u$ and $\omega$) are involved in $\zeta(\cdot)$,  we call the functional gradient $\zeta(\cdot)$ as quadruply stochastic functional gradient.

Then, we first give the update rule with the stochastic gradient $\xi(\cdot)$ as follow,
\begin{align}
h_{t+1}(\cdot) = h_{t}(\cdot)-\eta_t\left(\xi_t(\cdot)+\lambda h_t(\cdot)\right)= \sum_{i=1}^{t}a_t^i\xi_t(\cdot), \ \  \forall t>1, \nonumber
\end{align}
where $\mathit{a}_t^i=-\eta_t \prod_{j=i+1}^t(1-\eta_j\lambda)$, $\eta_t$ denotes the step size and $h_{t+1}(x)$ denotes the function value if we use gradient $\xi(\cdot)$. Since $\zeta(\cdot)$ is an unbiased estimate of $\xi(\cdot)$, the update rule using $\zeta(\cdot)$ after $t$ iterations can be  written as follow,
\begin{align}
f_{t+1}(\cdot) &= f_t(\cdot)-\eta_t\left (\zeta_t(\cdot)+\lambda f_t(\cdot)\right )=\sum_{i=1}^t a_t^i\zeta_i(\cdot), \quad \forall t > 1,\nonumber
\end{align}
where $f_1(\cdot) = 0$, and $f_{t+1}(x)$ denotes the function value for the input $x$ if we use the functional gradient $\zeta(\cdot)$.

In order to implement the update process in computer program, we rewrite the update rule as the following iterative update rules with constantly-changing coefficients $\{\alpha_i\}_{i=1}^t$,
\begin{eqnarray}
f_{t} &=& \sum_{i=1}^{t}\alpha_i\phi_{\omega}(x),
\end{eqnarray}
\begin{eqnarray}\label{update_alpha_i}
\alpha_i &=& -\eta_i (\gamma( l_1'\phi_{\omega}(x^p)+l_2'\phi_{\omega}(x^n))\nonumber\\&&+(1-\gamma)(l_3'\phi_{\omega}(x^p) +l_4'\phi_{\omega}(x^u)\nonumber\\&&+l_5'\phi_{\omega}(x^u)+l_6'\phi_{\omega}(x^n))),
\end{eqnarray}
\begin{eqnarray}\label{update_alpha_j}
\alpha_j &=& (1-\eta_j\lambda)\alpha_j, \;\mathrm{for}\; j=1,...,i-1.
\end{eqnarray}  

\begin{algorithm}[!ht]	
	\caption{$\{\alpha_i\}_{i=1}^t$ =\textbf{ QSG-S2AUC}$(D_p,D_n,p(x))$} 
	\renewcommand{\algorithmicrequire}{\textbf{Input:}}
	\renewcommand{\algorithmicensure}{\textbf{Output:}}
	\begin{algorithmic}[1] 
		\REQUIRE $p(\omega)$, $\phi_{\omega}(x)$, $l(u,v)$, $ \lambda$.
		\ENSURE $\{\alpha_i\}_{i=1}^t$
		\FOR{$i=1,...,t$}
		\STATE Sample $x^p$ from $D_p$.
		\STATE Sample $x^n$ from $D_n$.
		\STATE Sample $x^u \sim p(x)$.
		\STATE Sample $\omega_i \sim p(\omega)$ with seed $i$.
		\STATE $f(x_i)=$ \textbf{Predict}$(x_i,\{\alpha_i\}_{j=1}^{i-1})$.
		\STATE $\alpha_i = -\eta_i (\gamma( l_1'\phi_{\omega}(x^p)+l_2'\phi_{\omega}(x^n))+(1-\gamma)(l_3'\phi_{\omega}(x^p) +l_4'\phi_{\omega}(x^u)+l_5'\phi_{\omega}(x^u)+l_6'\phi_{\omega}(x^n)))$
		\STATE $\alpha_j = (1-\eta_j\lambda)\alpha_j \;for\; j=1,...,i-1$.
		\ENDFOR
	\end{algorithmic}
	\label{alg:train}
\end{algorithm}

\begin{algorithm}[!ht]
	\caption{$f(x)=$\textbf{Predict}$(x,\{\alpha_i\}_{i=1}^t)$} 
	\renewcommand{\algorithmicrequire}{\textbf{Input:}}
	\renewcommand{\algorithmicensure}{\textbf{Output:}}
	\begin{algorithmic}[1] 
		\REQUIRE $p(\omega),\phi_{\omega}(x)$
		\ENSURE $f(x)$
		\STATE Set $f(x) = 0$.
		\FOR{$i=1,...,t$}
		\STATE Sample $\omega_i \sim p(\omega)$ with seed $i$.
		\STATE $f(x)=f(x)+\alpha_i\phi_{\omega}(x)$
		\ENDFOR
	\end{algorithmic}
	\label{alg:predict}
\end{algorithm}

\subsection{QSG-S2AUC Algorithms}
In our implementation, we use \emph{pseudo-random number generators} with seed $i$ to sample random features. In each iteration, we only need to keep the seed $i$ aligned between prediction and training. Then the prediction function $f(x)$ can be restored much more easily. Besides, the QSG-S2AUC maintains a sequence of $\{\alpha_i\}_{i=1}^t$ at each ieration which has low memory requirement. Specifically, each iteration of the training algorithm executes the following steps.
\begin{enumerate}

\item \textit{Select Random Data Triplets:} \ Randomly sample a positive instance, a negative instance and an unlabeled instance to compose a data triplet. In addition, we use mini-batch of these data points to achieve a better efficiency.

\item \textit{Approximate the Kernel Function:} \ Sample $\omega_i \sim p(\omega)$ with random seed $i$ to calculate the random features on-the-fly. We keep this seed aligned between prediction and training to speed up computing $f(x_i) =\sum_{i=1}^t\alpha_i \phi_{\omega}(x)$.

\item \textit{Update Coefficients:} \  We compute the current coefficient $\alpha_i$ in $i$-th loop and then update the former coefficients $\alpha_j$ for $j=1,\cdots,i-1$ according to the update rule (\ref{update_alpha_i}) and (\ref{update_alpha_j}), respectively.

\end{enumerate}
We summarize the  algorithms for training and prediction in Algorithm \ref{alg:train} and \ref{alg:predict} respectively.

\section{Convergence Analysis}

In this section, we prove that QSG-S2AUC  converges to the optimal solution at the rate of $O(1/t)$. We first give several assumptions which are standard in DSG ~\cite{dai2014scalable}.
\begin{assumption}
	(Bound of kernel function). The kernel function is bounded, \textit{i.e.}, $k(x,x') \leq \kappa$, where $\kappa > 0$.
\end{assumption}
\begin{assumption}
	(Bound of random feature norm). The random feature norms are bounded, \textit{i.e.}, $|\phi_{\omega}(x)\phi_{\omega}(x')| \leq \phi$.
\end{assumption}
\begin{assumption}\label{assumption:lipchitz_continuous}
	(Lipschitz continuous). The first order derivation of $l(f(x^p),f(x^n))$ is $L_1$-\textbf{Lipschitz continuous} in terms of $f(x^p)$ and $L_2$-\textbf{Lipschitz continuous} in terms of $f(x^n)$. Similarly, the first order derivation of $l(f(x^p),f(x^u))$ is $L_3$-\textbf{Lipschitz continuous} in terms of $f(x^p)$ and $L_4$-\textbf{Lipschitz continuous} in terms of $f(x^u)$ and the first order derivation of $l(f(x^u),f(x^n))$ is $L_5$-\textbf{Lipschitz continuous} in terms of $f(x^u)$ and $L_6$-\textbf{Lipschitz continuous} in terms of $f(x^n)$.
\end{assumption}

\begin{assumption}
	(Bound of derivation). There exists $M_1>0$, $M_2>0$, $M_3>0$, $M_4>0$, $M_5>0$ and $M_6>0$, such that $|l_1'| \leq M_1$, $|l_2'| \leq M_2$, $|l_3'| \leq M_3$, $|l_4'| \leq M_4$, $|l_5'| \leq M_5$, $|l_6'| \leq M_6$,
\end{assumption}

We use the  framework of ~\cite{dai2014scalable} to prove that $f_{t+1}$ can converge to the optimal solution $f^*$. Specifically, we use the aforementioned $h_{t+1}$ as an intermediate value to decompose the difference between $f_{t+1}$ and $f^{*}$ as follows,
\begin{eqnarray}\label{total_err}
&&|f_{t+1}(x) - f^{*}(x)|^2 \nonumber\\&\leq& 2 \underbrace{|f_{t+1}(x)-h_{t+1}(x)|^2}_{\rm error \; due\; to\; random\; features}  + 2\kappa \underbrace{\| h_{t+1} -f^{*} \|_{\mathcal{H}}.}_{\rm error\; due\; to\; random\; data}.
\end{eqnarray}
In other words, the total approximation error includes the error caused by approximating the kernel with random features, and the error caused by sampling random data. Finally, the boundary of the original error can be obtained by summing up the boundary of these two parts. 

We first give the convergence of error due to random features and random data in Lemmas \ref{lemma:error_due_random_feature} and \ref{lemma:error_due_to_random_data} respectively. All the detailed proofs are provided in our Appendix. 
\begin{lemma}[Error due to random features]\label{lemma:error_due_random_feature}
	Let $\chi$ denotes the whole training set in semi-supervised learning problem. For any $x \in \chi$, we have
	\begin{eqnarray}
	\mathbb{E}_{x_t^p,x_t^n,x_t^u,\omega}[|f_{t+1}(x)-h_{t+1}(x)|^2] \leq B^2_{1,t+1},
	\end{eqnarray}
	where $B^2_{1,t+1}:=M^2(\kappa+\phi)^2\sum_{i=1}^{t}|a_t^i|^2$, $B_{1,1} = 0$ and $M = \gamma(M_1+M_2)+(1-\gamma)(M_3+M_4+M_5+M_6)$.
\end{lemma}
Obviously, the upper bound $B^2_{1,t+1}$ depends on the convergence of $|a_t^i|$, which is given in Lemma \ref{lemma:upper_bound}.
\begin{lemma}\label{lemma:upper_bound}		
	Suppose $\eta_i = \dfrac{\theta}{i}$ ($1 \leq  i \leq t$) and $\theta\lambda \in (1,2)\cup\mathbb{Z}_{+}$. We have $|a_t^i| \leq \dfrac{\theta}{t}$ and $\sum_{i=1}^t |a_t^i|^2 \leq \dfrac{\theta^2}{t}$.
\end{lemma}
\begin{figure*}[!t]
	\begin{subfigure}[b]{0.24\textwidth}
		\centering
		\includegraphics[width=1.8in]{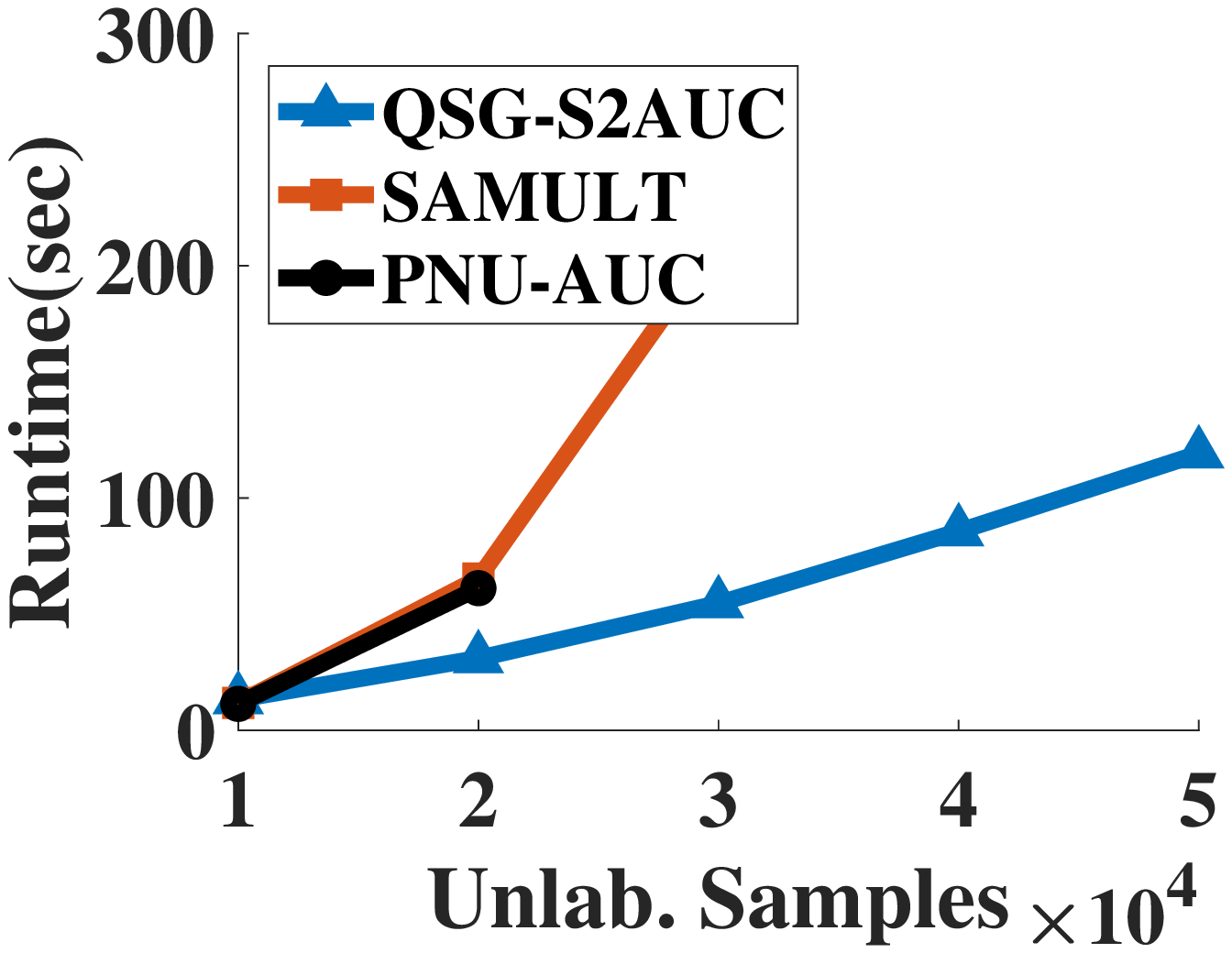}
		\caption{CodRNA}
	\end{subfigure}
	\begin{subfigure}[b]{0.24\textwidth}
		\centering
		\includegraphics[width=1.8in]{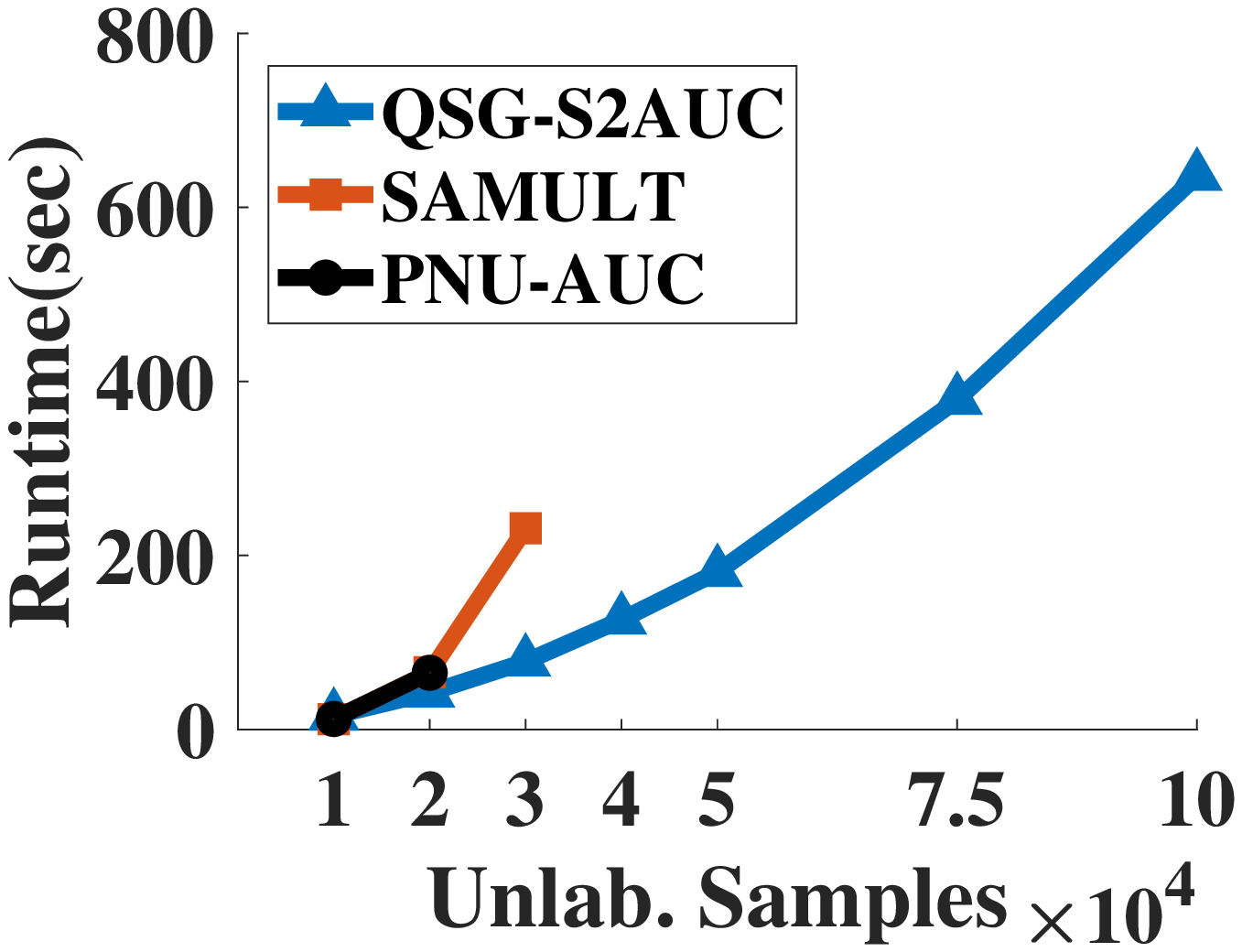}
		\caption{Covtype}
	\end{subfigure}
	\begin{subfigure}[b]{0.24\textwidth}
		\centering
		\includegraphics[width=1.8in]{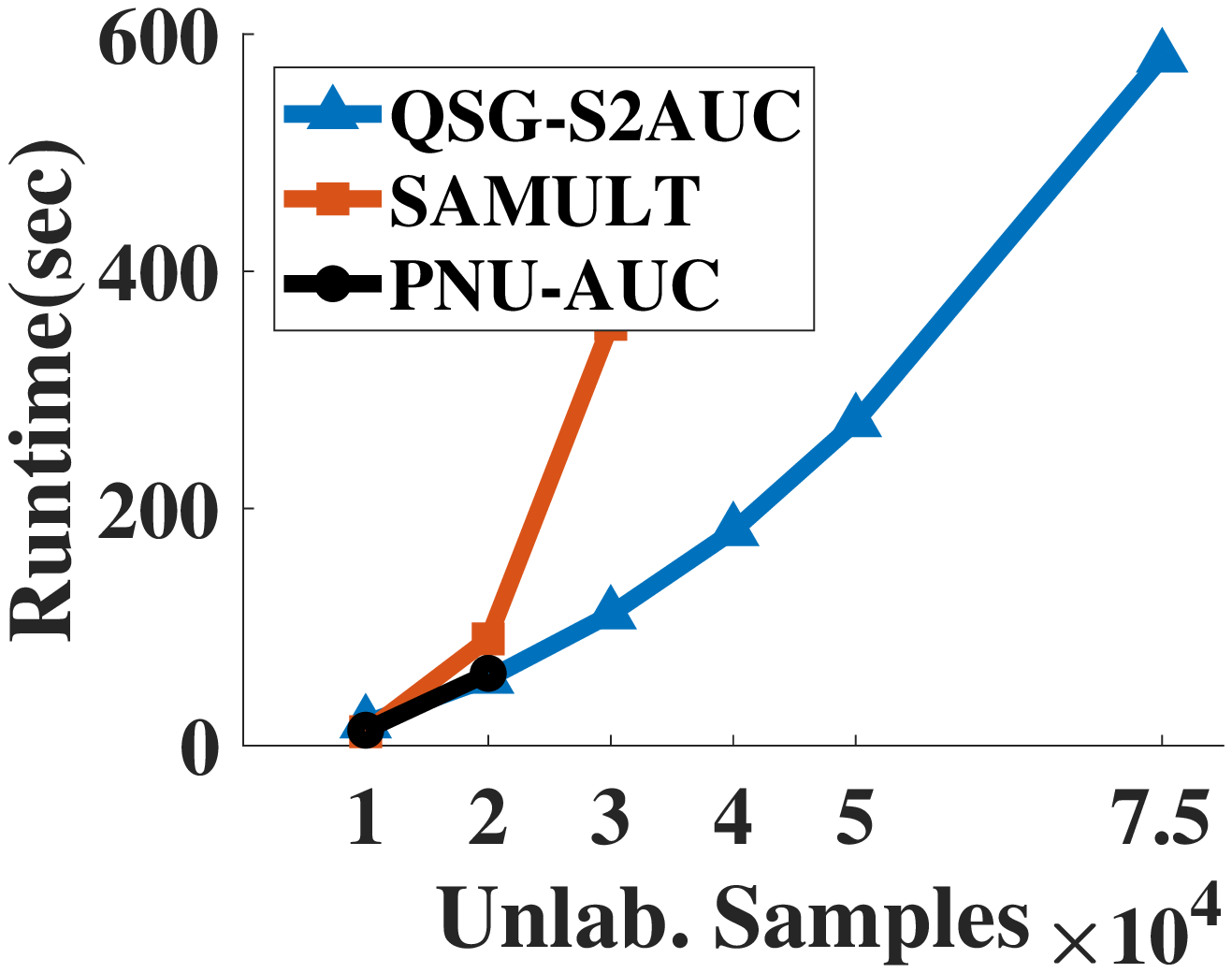}
		\caption{Dota2}
	\end{subfigure}	
	\begin{subfigure}[b]{0.24\textwidth}
		\centering
		\includegraphics[width=1.8in]{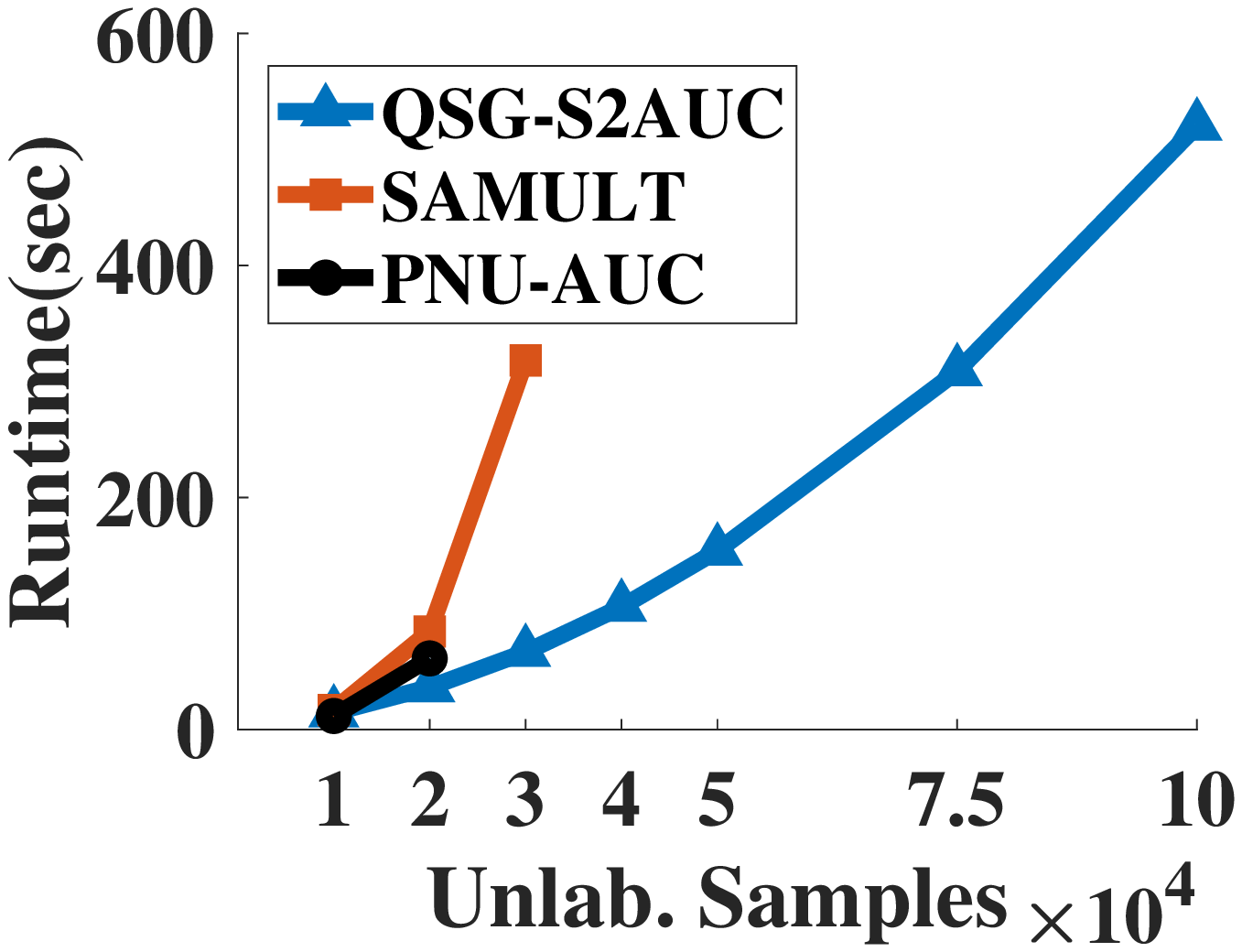}
		\caption{Higgs}
	\end{subfigure}

	\begin{subfigure}[b]{0.24\textwidth}
		\centering
		\includegraphics[width=1.8in]{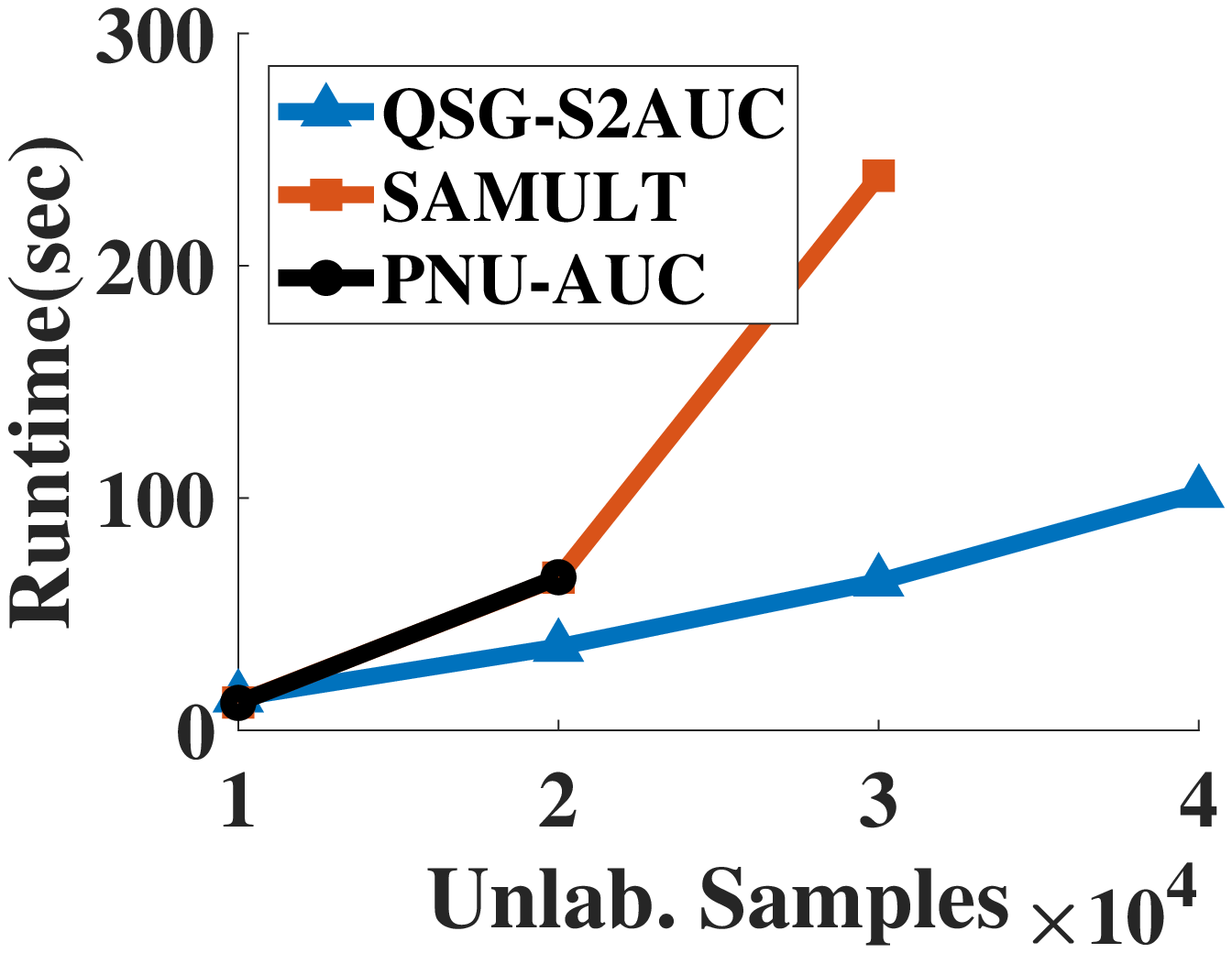}
		\caption{IJCNN1}
	\end{subfigure}	
	\begin{subfigure}[b]{0.24\textwidth}
		\centering
		\includegraphics[width=1.8in]{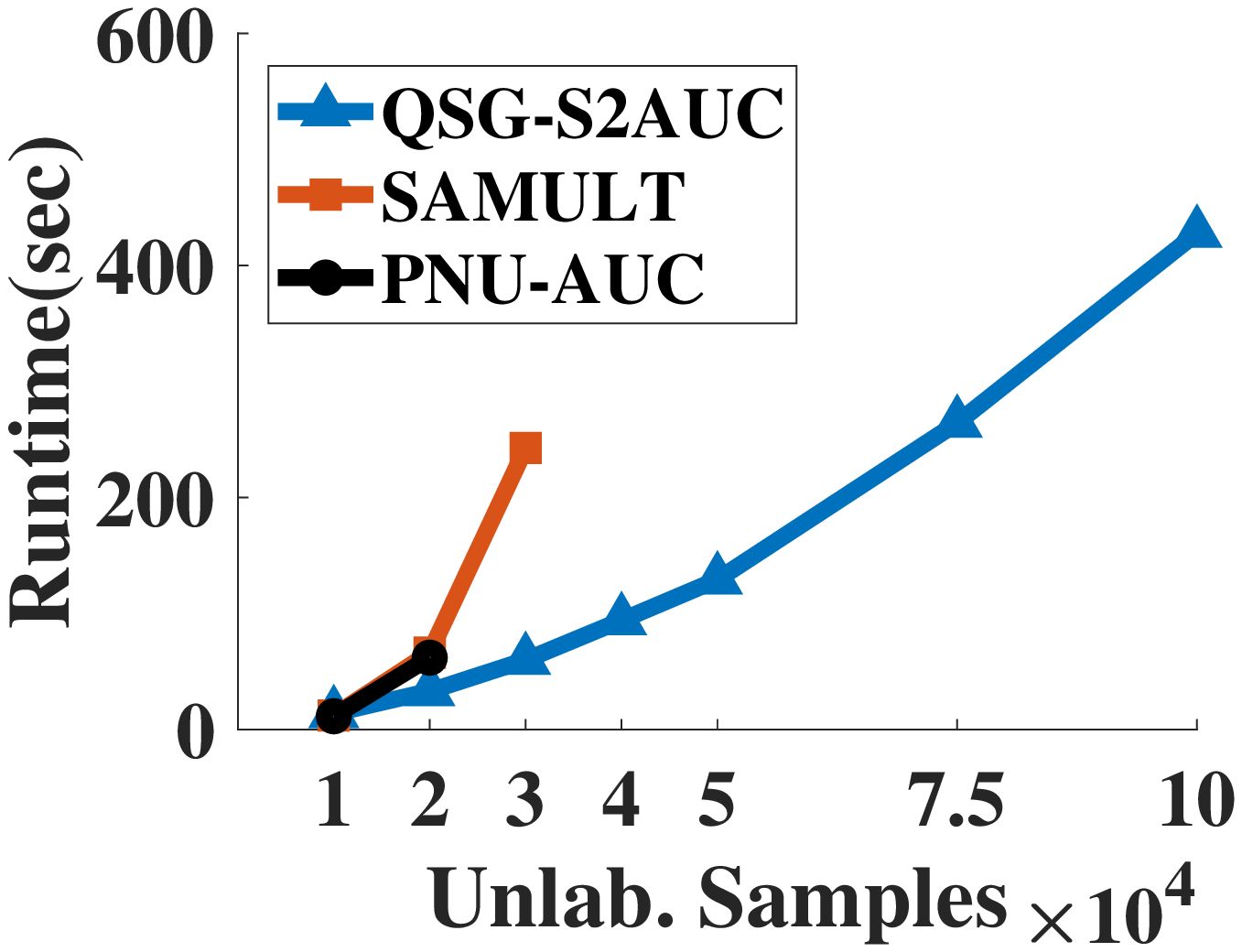}
		\caption{Skin}
	\end{subfigure}	
	\begin{subfigure}[b]{0.24\textwidth}
		\centering
		\includegraphics[width=1.8in]{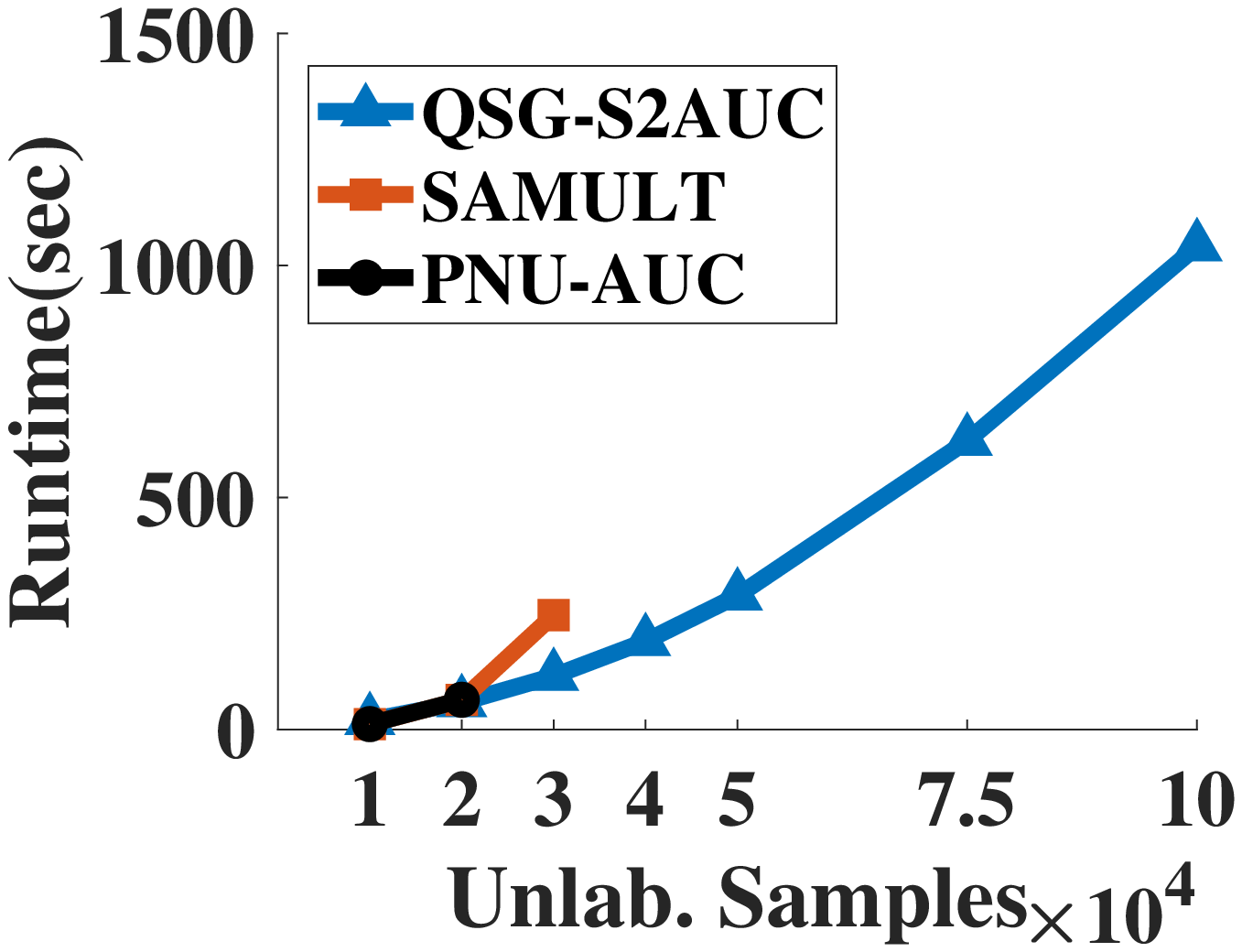}
		\caption{Unclonable}
	\end{subfigure}
	\begin{subfigure}[b]{0.24\textwidth}
		\centering
		\includegraphics[width=1.8in]{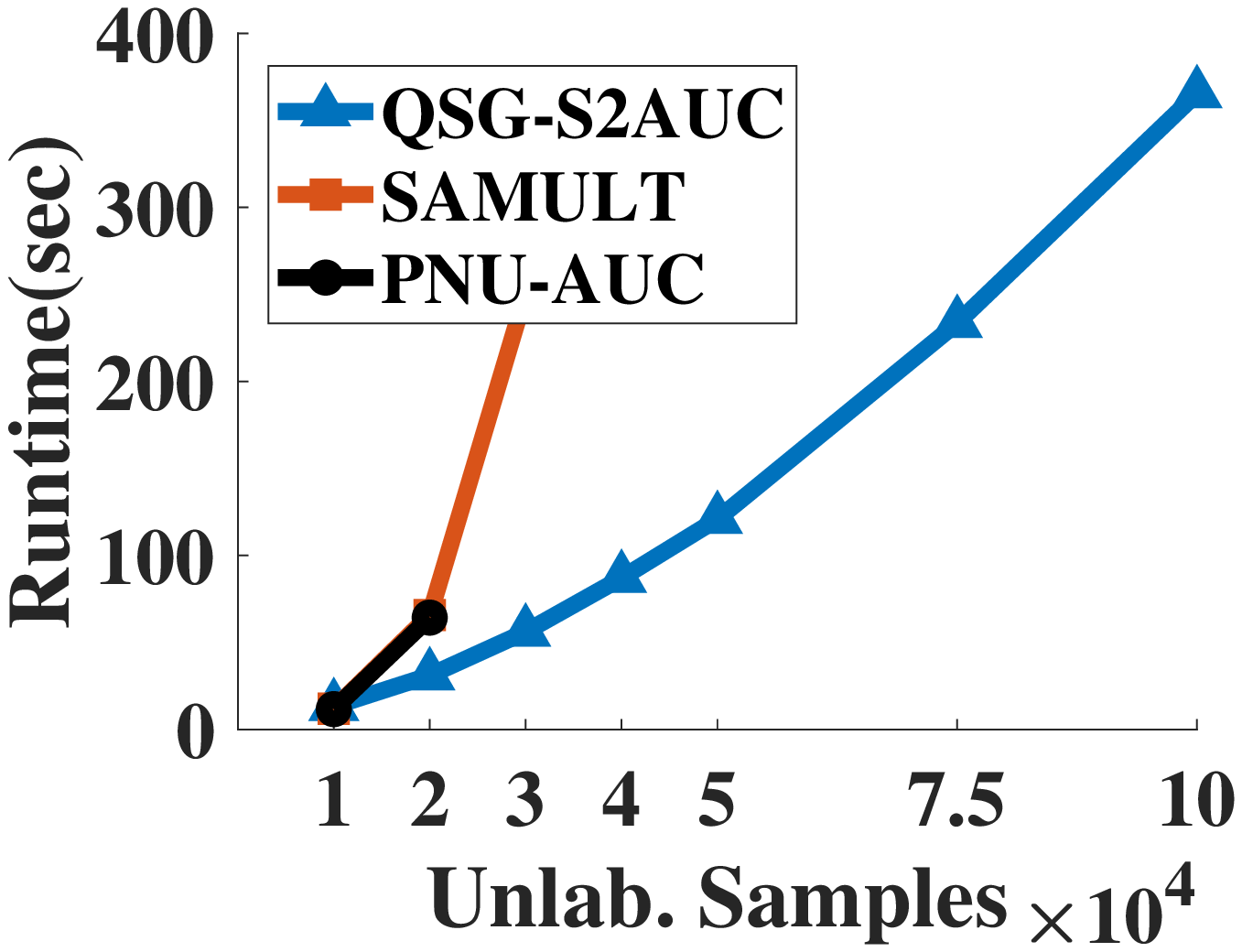}
		\caption{SUSY}
	\end{subfigure}			
	\caption{The  training time of QSG-S2AUC, SAMULT and PNU-AUC  against different sizes of unlabeled samples, where the sizes of  labeled samples are fixed at 200. (The lines of SAMULT and PNU-AUC are incomplete  because their implementations crash on  larger training sets.)}
	\label{fig:time_vs_unlabeled}
\end{figure*}
\begin{table}	
	\small
	\centering
	\setlength{\tabcolsep}{3mm}
	\begin{tabular}{cccc}
		\toprule
		\textbf{Dataset}  & \textbf{Features} & \textbf{Samples} & \textbf{Source} \\ \hline		
		Codrna  &       8        &  59,535 &LIBSVM\\
		Ijcnn1   &       22       &  49,990 & LIBSVM \\
		Susy    &      18      &  5,000,000 &LIBSVM \\ 		
		Covtype 	& 54& 581,012 & LIBSVM\\
		Higgs &  28& 1,100,000 &LIBSVM \\
		Skin  &3& 245,057 &		LIBSVM\\
		\hline
		Dota2  &116&92650 &	UCI\\
		Unclonable &129&6,000,000 & UCI\\
		\bottomrule
	\end{tabular}
	
	\caption{Datasets used in the experiments.}
	\label{tab:dataset}
	
\end{table}
\begin{figure}[!ht]
	\centering	
	\includegraphics[scale=0.45]{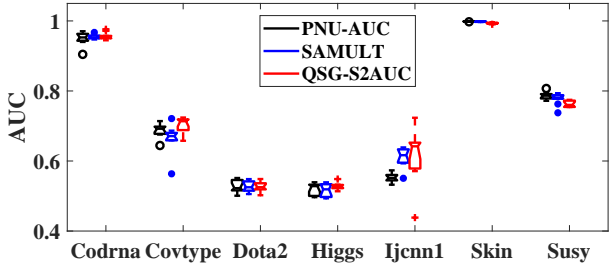}	
	\caption{The boxplot of testing AUC results for  PNU-AUC, SAMULT and our QSG-S2AUC.}
	\label{fig:auc_result}
\end{figure}

\begin{remark}
	According to  Lemmas \ref{lemma:error_due_random_feature}  and \ref{lemma:upper_bound}, the error caused by random features has the convergence rate of $O(1/t)$ with proper learning rate and $\theta\lambda \in (1,2)$.
\end{remark}

\begin{lemma}[Error due to random data]\label{lemma:error_due_to_random_data}
	Set $\eta_t = \dfrac{\theta}{t}$, $\theta>0$, such that $\theta\lambda \in (1,2)\cup \mathbb{Z}_{+}$, we have
	\begin{equation}		
	\mathbb{E}_{x_t^p,x_t^n,x_t^u,\omega_t}\left[\| h_{t+1} - f^{*} \|_{\mathcal{H}}^2\right] \leq \dfrac{Q_1^2}{t},		
	\end{equation}
	where
	$Q_1 = \max\left\{\| f^{*} \|_{\mathcal{H}},\frac{Q_0 + \sqrt{Q_0^2+(2\theta\lambda-1)(1+\theta\lambda)^2\theta^2\kappa M^2}}{2\theta\lambda-1} \right\} $,
	$Q_0 = \sqrt{2}\kappa^{1/2}(\kappa+\phi)LM\theta^2$ and $L = \gamma(L_1+L_2)+(1-\gamma)(L_3+L_4+L_5+L_6)$.
\end{lemma}
According to Lemmas \ref{lemma:error_due_random_feature} and  \ref{lemma:error_due_to_random_data}, we can obtain the convergence rate of QSG-S2AUC  in Theorem \ref{theorem:Convergence_in_expectation}.
\begin{theorem}[Convergence in expectation]\label{theorem:Convergence_in_expectation}
	Let $\chi$ denote the whole training set in semi-supervised learning problem. Set $\eta_t = \dfrac{\theta}{t}$, $\theta >0$, such that $\theta\lambda \in (1,2)\cup \mathbb{Z}_{+}$. $\forall x \in \chi$, we have
	\begin{align}
	\mathbb{E}_{x_t^p, x_t^n, x_t^u,\omega_t}\left[|f_{t+1}(x)-f^{*}(x)|^2\right] \leq \dfrac{2C^2+2\kappa Q^2_1}{t},  \nonumber
	\end{align}
	where $C^2 = (\kappa+ \phi)^2M^2\theta^2$.
\end{theorem}
\begin{remark}
	Theorem \ref{theorem:Convergence_in_expectation} shows that for any given $x$, the evaluated value of $f_{t+1}$ at $x$ will converge to that of $f^{*}$ in terms of the Euclidean distance at the rate of $O(1/t)$. This rate is the same as that of standard DSG even though our problem is much more complicated and has four sources of randomness.
\end{remark}

\section{Experiments}
In this section, we present the experimental results on several datasets to demonstrate the effectiveness and efficiency of QSG-S2AUC.

\subsection{Experimental Setup}
We compare the AUC
results and running time of QSG-S2AUC with the state-of-the-art semi-supervised AUC maximization algorithms as summarized as follows.
\begin{enumerate}

\item \textit{\textbf{PNU-AUC}}: Unbiased semi-supervised AUC optimization method proposed in ~\cite{sakai2018semi} based on positive and unlabeled learning.

\item \textit{\textbf{SAMULT}}: The method proposed in ~\cite{xie2018semi} which does not require the class prior distribution to achieve the unbiased solution.

\end{enumerate}

All the experiments were ran on a PC with 56 2.2GHz cores and 80GB RAM. We implemented  QSG-S2AUC and SAMULT algorithms in MATLAB. We used the MATLAB code from \url{https://github.com/t-sakai-kure/PNU} as the implementation of PNU-AUC. For all algorithms, we use the square pairwise loss $l(u,v)=(1-u+v)^2$  and Gaussian  kernel $k(x,x') = \exp(-\sigma \| x-x' \| ^2)$. The hyper-parameters ($\lambda$, $\sigma$ and $\gamma$) are chosen via 5-fold  cross-validation. 
$\lambda$ and $\sigma$ were searched in the region $\{ (\lambda , \sigma)|  2^{-3} \le  \lambda \le 2^3,\; 2^{-3} \le  \sigma \le 2^3\}$. The trade-off parameter $\gamma$ in SAMULT and QSG-S2AUC was searched from $0$ to $1$ at intervals of $0.1$, and that in PNU-AUC was searched from $-1$ to $1$ at intervals of $0.1$. In addition, the class prior $\pi$ in PNU-AUC is set to the class proportion in the whole training set, which can be estimated by ~\cite{du2015class}.
All the results are the average of 10 trials.

\subsection{Datasets}
We carry out the experiments on eight large scale benchmark datasets collected from LIBSVM\footnote{LIBSVM is available at \url{https://www.csie.ntu.edu.tw/~cjlin/libsvmtools/\\datasets/binary/}.} and UCI\footnote{UCI is available at  \url{http://archive.ics.uci.edu/ml/datasets.html}.} repositories. The size $n$ of the dataset and the feature dimensionality $d$ are summarized in Table \ref{tab:dataset}. To conduct the experiments for semi-supervised learning, we randomly sample 200 labeled instances and treat the rest of the data as unlabeled. All the data features are normalized to $[0,1]$ in advance.

\subsection{Results and Discussion}
Figure \ref{fig:time_vs_unlabeled} shows the training time of the three algorithms against different sizes of unlabeled samples on the eight benchmark datasets, where the sizes of labeled samples are fixed at 200. We can find that QSG-S2AUC is always faster than SAMULT and PNU-AUC. This is because the SAMULT and PNU-AUC need $O(n^3)$ operations to compute the inverse matrixes with kernel. Differently, QSG-S2AUC uses RFF to approximate the kernel function, and each time it only needs $O(D)$ operations to calculate the random features with seed $i$. In addition, the low memory requirement of QSG-S2AUC allows it to do an efficient training for large scale datasets while PNU-AUC and SAMULT are out of memory. Figure \ref{fig:auc_result} presents the testing AUC results of these algorithms on the eight benchmark datasets. The results show that QSG-S2AUC has the similar AUC results with other methods on the most datasets, and  has the highest AUC on the datasets of Covtype and Ijcnn1. Based on these results, we conclude that QSG-S2AUC is superior to other state-of-the-art algorithms in terms of efficiency and scalability, while retaining the similar generalization performance.


\section{Conclusion}
In this paper, we propose a novel scalable semi-supervised AUC optimization algorithm, QSG-S2AUC. Considering that semi-supervised learning contains three data sources, DSG-S2AUC is designed to randomly sample one instance from each data source in each iteration. Then, their random features are generated and used to calculate a quadruply stochastic functional gradient for model update. Even though this optimization process contains multiple layers of stochastic sampling, theoretically, we prove that QSG-S2AUC has a convergence rate of $O(1/t)$. The experimental results on various datasets also demonstrate the superiority of the proposed QSG-S2AUC. 
\section*{Acknowledgments}
H.H. was partially supported by U.S. NSF IIS 1836945, IIS 1836938, DBI 1836866, IIS 1845666, IIS 1852606, IIS 1838627, IIS 1837956. B.G. was partially supported by the National Natural Science Foundation of China (No: 61573191), and the Natural Science Foundation  (No. BK20161534), Six talent peaks project (No. XYDXX-042) in Jiangsu Province.

\bibliographystyle{named}
\bibliography{ijcai19}

\appendix
\section{Detailed Proof of Convergence Rate}
In this section, we give detailed proof of Lemmas \ref{lemma:error_due_random_feature}-\ref{lemma:error_due_to_random_data} and Theorem \ref{theorem:Convergence_in_expectation}.

\subsection{Proof of Lemma \ref{lemma:error_due_random_feature}}

Here we give the detailed proof of Lemma \ref{lemma:error_due_random_feature}.
\begin{proof}
	We denote $A_i(x)=A_i(x;x_i^p,x_i^n,x_i^u,\omega_i) := a_t^i(\zeta_i(x)-\xi_i(x))$. According to the assumption in section 5, $A_i(X)$ have a bound:	
	\begin{eqnarray}
	|A_i(x)| &\leq& |a_t^i|\left(|\zeta_i(x)|+|\xi_i(x)|\right)  \nonumber\\
	&\leq&|a_t^i|(\gamma(l'_1k(x_i^p,x)+l'_2k(x_i^n,x))\nonumber\\&&+(1-\gamma)(l'_3k(x_i^p,x)+l'_4k(x_i^u,x)\nonumber\\&&+l'_5k(x_i^u,x)+l'_6k(x_i^n,x))\nonumber \\&&+ \gamma(l_1'\phi_{\omega}(x^p)\phi_{\omega}(x)+l_2'\phi_{\omega}(x^n)\phi_{\omega}(x)\nonumber\\&&+(1-\gamma)(l_3'\phi_{\omega}(x^p)\phi_{\omega}(x)+l_4'\phi_{\omega}(x^u)\phi_{\omega}(x)\nonumber\\&&+l_5'\phi_{\omega}(x^u)\phi_{\omega}(x)+l_6'\phi_{\omega}(x^n)\phi_{\omega}(x)))) \nonumber\\	
	&\leq&|a_t^i|((\gamma(M_1+M_2)\nonumber\\&&+(1-\gamma)(M_3+M_4+M_5+M_6))\kappa\nonumber\\&&+(\gamma(M_1+M_2)\nonumber\\&&+(1-\gamma)(M_3+M_4+M_5+M_6))\phi)\nonumber\\
	&=& M(\kappa+\phi)|a_t^i | \nonumber
	\end{eqnarray}
	
	Then we obtain the lemma \ref{lemma:error_due_random_feature}. This completes the proof.
\end{proof}

\subsection{Proof of lemma \ref{lemma:upper_bound}}
Here we give detailed proof of Lemma \ref{lemma:upper_bound}.
\begin{proof}
	Obviously, $|a_t^i| \leq \dfrac{\theta}{t}$. Then we have
	\begin{align}
	|a_t^i| &= |a_t^{i+1}\dfrac{\eta_i}{\eta_{i+1}}(1-\lambda\eta_{i+1})|\nonumber\\
	&= \dfrac{i+1}{i}|1-\dfrac{\lambda\theta}{i+1}|\cdot|a_t^{i+1}|\nonumber\\
	&= |\dfrac{i+1-\lambda\theta}{i}|\cdot|a_t^{i+1}| \nonumber
	\end{align}
	When $\lambda\theta \in (1,2), \forall i \geq 1$, we have $i-1<i+1-\lambda\theta < i$, so $|a_t^{i}|<|a_t^{i+1}|\leq \dfrac{\theta}{t}$ and $\sum_{i=1}^{t} \leq \dfrac{\theta^2}{t}$. When $\lambda\theta \in \mathbb{Z}_{+}$, if $i> \lambda\theta-1$, then $|a_t^i|< |a_t^{i+1}|\leq \dfrac{\theta}{t}$. If $i \leq \lambda\theta -1$, then $|a_t^i|=0$. So we get $\sum_{i=1}^t|a_t^i|^2 \leq \dfrac{\theta^2}{t}$.
	Therefore, we obtain the lemma \ref{lemma:upper_bound}. This completes the proof.
\end{proof}

\subsection{Proof of Lemma \ref{lemma:error_due_to_random_data}}

For convenience, we denote that $l_1'(f(x_t^p),f(x_t^n))$, $l_2'(f(x_t^p),f(x_t^n))$, $l_3'(f(x_t^p),f(x_t^u))$, $l_4'(f(x_t^p),f(x_t^u))$, $l_5'(f(x_t^u),f(x_t^n))$ and $l_6'(f(x_t^u),f(x_t^n))$ as $l_1'(f_t)$, $l_2'(f_t)$, $l_3'(f_t)$, $l_4'(f_t)$, $l_5'(f_t)$ and $l_6'(f_t)$, respectively. In addition, we define the following three different gradient terms,
\begin{eqnarray}
g_t &=& \xi_t + \lambda h_t\nonumber\\ 
&=&\gamma(l_1'(f_t)k(x_t^p,\cdot)+l_2'(f_t)k(x_t^n,\cdot))\nonumber\\&&+(1-\gamma)(l_3'(f_t)k(x_t^p,\cdot)+l_4'(f_t)k(x_t^u,\cdot)\nonumber\\&&+l_5'(f_t)k(x_t^u,\cdot)+l_6'(f_t)k(x_t^n,\cdot))+\lambda h_t \nonumber 
\end{eqnarray}
\begin{eqnarray}
\hat{g_t} &=&\hat{\xi_t} + \lambda h_t \nonumber\\
&=&\gamma(l_1'(h_t)k(x_t^p,\cdot)+l_2'(h_t)k(x_t^n,\cdot))\nonumber\\&&+(1-\gamma)(l_3'(h_t)k(x_t^p,\cdot)+l_4'(h_t)k(x_t^u,\cdot)\nonumber\\&&+l_5'(h_t)k(x_t^u,\cdot)+l_6'(h_t)k(x_t^n,\cdot))+\lambda h_t \nonumber 
\end{eqnarray}
\begin{eqnarray}
\bar{g_t}&=& \mathbb{E}_{x_t^p,x_t^n,x_t^u}[\hat{g_t}]\nonumber\\
&=&\mathbb{E}_{x_t^p,x_t^nx_t^u}[\gamma(l_1'(h_t)k(x_t^p,\cdot)+l_2'(h_t)k(x_t^n,\cdot))\nonumber\\&&+(1-\gamma)(l_3'(h_t)k(x_t^p,\cdot)+l_4'(h_t)k(x_t^u,\cdot)\nonumber\\&&+l_5'(h_t)k(x_t^u,\cdot)+l_6'(h_t)k(x_t^n,\cdot))]+\lambda h_t \nonumber
\end{eqnarray}

Note that from our previous definition, we have $h_{t+1}=h_t-\eta_t g_t, \forall t\geq 1$.

Denote $A_t=\parallel h_t - f^{*}\parallel_{\mathcal{H}}^2$. Then we have
\begin{eqnarray}
A_{t+1} &=& \parallel h_t - f^{*} - \eta_t g_t \parallel_{\mathcal{H}}^2 \nonumber \\
&=& A_t+ \eta_t^2 \parallel g_t \parallel_{\mathcal{H}}^2 - 2\eta_t \langle h_t - f^{*},g_t \rangle_{\mathcal{H}} \nonumber \\
&=& A_t + \eta_t^2 \parallel g_t \parallel_{\mathcal{H}}^2-2\eta_t \langle h_t - f^{*}, \bar{g_t} \rangle_{\mathcal{H}} \nonumber \\&& + 2\eta_t\langle h_t-f^{*}, \bar{g_t}-\hat{g_t} \rangle_{\mathcal{H}} + 2\eta_t \langle h_t-f^{*}, \hat{g_t}-g_t \rangle_{\mathcal{H}} \nonumber
\end{eqnarray}
Because of the strongly convexity of loss function and optimality condition, we have
\begin{eqnarray}
\langle h_t - f^{*},\bar{g_t} \rangle_{\mathcal{H}} \geq \lambda \parallel h_t - f^{*} \parallel_{\mathcal{H}}^2 \nonumber
\end{eqnarray}

Hence, we have
\begin{align}\label{random_data_err_function}
A_{t+1} &\leq (1-2\eta_t\lambda)A_t + \eta_t^2 \parallel g_t \parallel_{\mathcal{H}}^2 + 2\eta_t\langle h_t-f^{*},\bar{g_t}- \hat{g_t}  \rangle_{\mathcal{H}} \nonumber\\
& \quad + 2\eta_t \langle h_t-f^{*},\hat{g_t}-g_t \rangle_{\mathcal{H}}, \forall t \geq 1
\end{align}
Let us denote $\mathcal{M}_t = \parallel g_t \parallel_{\mathcal{H}}^2$, $\mathcal{N}_t = \langle h_t-f^{*},\bar{g_t}-\hat{g_t} \rangle_{\mathcal{H}}  $, $\mathcal{R}_t = \langle h_t-f^{*},\hat{g_t}-g_t\rangle_{\mathcal{H}} $. Firstly, we show that $\mathcal{M}_t$, $\mathcal{N}_t$, $\mathcal{R}_t$ are bounded. Specifically, for $t\geq 1$, we have
\begin{align}\label{Eq_M}
\mathcal{M}_t \leq \kappa M^2 (1+\lambda c_t)
\end{align}
where $c_t := \sqrt{\sum_{i,j=1}^{t-1}|a_{t-1}^{i}||a_{t-1}^{j}|}$
\begin{align}\label{Eq_N}
\mathbb{E}_{x_t^p,x_t^n,x_t^u,\omega_t}[\mathcal{N}_t] = 0
\end{align}
\begin{align}\label{Eq_R}
\mathbb{E}_{x_t^p,x_t^nx_t^u,\omega_t}[\mathcal{R}_t] \leq \kappa^{1/2}LB_{1,t}\sqrt{\mathbb{E}_{x_{t-1}^p,x_{t-1}^n,x_{t-1}^u,\omega_{t-1}}[A_t]}
\end{align}
where $M:=\gamma(M_1+M_2)+(1-\gamma)(M_3+M_4+M_5+M_6)$, $L:=\gamma(L_1+L_2)+(1-\gamma)(L_3+L_4+L_5+L_6)$ and $A_t = \parallel h_t - f^{*}\parallel_{\mathcal{H}}^2$.

\begin{proof}
	The proof of Eq. (\ref{Eq_M}):
	
	\begin{align}
	\mathcal{M}_t = \parallel g_t \parallel_{\mathcal{H}}^2 = \parallel \xi_t+\lambda h_t \parallel_{\mathcal{H}} ^2 \leq \left(\parallel \xi \parallel_{\mathcal{H}} + \lambda \parallel h_t \parallel_{\mathcal{H}}\right)^2 \nonumber
	\end{align}
	and
	\begin{eqnarray}
	\parallel \xi_t \parallel_{\mathcal{H}} &=& \parallel \gamma (l_1'(f_t)k(x_t^p,\cdot)+l_2'(f_t)k(x_t^n,\cdot))\nonumber\\&&+(1-\gamma)(l_3'(f_t)k(x_t^p,\cdot) +l_4'(f_t)k(x_t^u,\cdot)\nonumber\\&&+ l_5'(f_t)k(x_t^u,\cdot)+l_6'(f_t)k(x_t^n,\cdot)) \parallel_{\mathcal{H}} \nonumber \\
	&\leq& \kappa^{1/2}(\gamma(M_1+M_2)\nonumber\\&&+(1-\gamma)(M_3+M_4+M_5+M_6)) \nonumber\\
	&=&\kappa^{1/2}M \nonumber
	\end{eqnarray}
	Then we have:
	\begin{eqnarray}
	\|h_t\|_{\mathcal{H}}^2&=&\sum_{i=1}^{t-1}\sum_{j=1}^{t-1}a_{t-1}^{i}a_{t-1}^{j}[\gamma(l_1'k(x_i^p,\cdot)+l_2'k(x_i^n,\cdot))\nonumber\\&&+(1-\gamma)(l_3'k(x_i^p,\cdot)+l_4'k(x_i^u,\cdot)\nonumber\\&&+l_5'k(x_i^u,\cdot)+l_6'k(x_i^n,\cdot))]\nonumber\\&&\cdot[\gamma(l_1'k(x_j^p,\cdot)+l_2'k(x_j^n,\cdot))\nonumber\\&&+(1-\gamma)(l_3'k(x_j^p,\cdot)+l_4'k(x_j^u,\cdot)\nonumber\\&&+l_5'k(x_j^u,\cdot)+l_6'k(x_j^n,\cdot))]\nonumber\\
	&\leq&\kappa\sum_{i=1}^{t-1}\sum_{j=1}^{t-1}a_{t-1}^{i}a_{t-1}^{j}[\gamma^2(M_1+M_2)^2\nonumber\\&&+2\gamma(1-\gamma)(M_1+M_2)(M_3+M_4+M_5+M_6)\nonumber\\&&+(1-\gamma)^2(M_3+M_4+M_5+M_6)^2]\nonumber \\
	&=&\kappa\sum_{i=1}^{t-1}\sum_{j=1}^{t-1}a_{t-1}^{i}a_{t-1}^{j}[\gamma(M_1+M_2)\nonumber\\&&+(1-\gamma)(M_3+M_4+M_5+M_6)]^2\nonumber\\
	&=& \kappa M^2\sum_{i=1}^{t-1}\sum_{j=1}^{t-1}a_{t-1}^{i}a_{t-1}^{j}  \nonumber
	\end{eqnarray}
	Then we obtain Eq. (\ref{Eq_M}). 
\end{proof}
\begin{proof}
	The proof of Eq. (\ref{Eq_N})
	\begin{eqnarray}
	&&\mathbb{E}_{x_t^p,x_t^n,x_t^u,\omega_t}[\mathcal{N}_t]\nonumber\\ 
	&=&\mathbb{E}_{x_{t-1}^p,x_{t-1}^n,x_{t-1}^u,\omega_t}\nonumber\\&&\left[\mathbb{E}_{x_{t}^p,x_t^n,x_{t}^u}[\langle h_t-f^{*},\bar{g_t}-\hat{g_t}\rangle_{\mathcal{H}}|x_{t-1}^p,x_{t-1}^u,\omega_t]\right] \nonumber\\
	&=& \mathbb{E}_{x_{t-1}^p,x_{t-1}^n,x_{t-1}^u,\omega_t}\left[\langle h_t-f^{*}, \mathbb{E}_{x_{t-1}^p,x_{t-1}^n,x_{t-1}^u}[\bar{g_t}-\hat{g_t}]\rangle_{\mathcal{H}}\right]\nonumber\\
	&=& 0 \nonumber
	\end{eqnarray}
\end{proof}

\begin{proof}
	The proof of Eq. (\ref{Eq_R})
	\begin{eqnarray}
	&& \mathbb{E}_{x_t^p,x_t^n,x_t^u,\omega_t}[\mathcal{R}_t]\nonumber\\	
	&=&\mathbb{E}_{x_{t}^p,x_t^n,x_{t}^u,\omega_t}[\langle h_t-f^{*},\hat{g_t}-g_t \rangle_{\mathcal{H}}] \nonumber\\	
	&=& \mathbb{E}_{x_{t}^p,x_t^n,x_{t}^u,\omega_t}[\langle h_t-f^{*}, \gamma[(l'_1(f_t)-l'_1(h_t))k(x_t^p,\cdot)\nonumber\\&&+(l'_2(f_t)-l'_2(h_t))k(x_t^n,\cdot)]\nonumber\\&&+(1-\gamma)[(l'_3(f_t)-l'_3(h_t))k(x_t^p,\cdot)\nonumber\\&&+(l'_4(f_t)-l'_4(h_t))k(x_t^u,\cdot)\nonumber\\&&+(l'_5(f_t)-l'_5(h_t))k(x_t^u,\cdot)\nonumber\\&&+(l'_6(f_t)-l'_6(h_t))k(x_t^n,\cdot)]  \rangle_{\mathcal{H}}] \nonumber\\
	&\leq &\mathbb{E}_{x_{t}^p,x_t^p,x_{t}^u,\omega_t} [\parallel h_t-f^{*} \parallel_{\mathcal{H}}\nonumber\\&&\cdot \gamma[|l'_1(f_t)-l'_1(h_t)|\cdot\parallel k(x_t^p,\cdot)\parallel_{\mathcal{H}}\nonumber\\&&+|l'_2(f_t)-l'_2(h_t)|\parallel k(x_t^n,\cdot)\parallel_{\mathcal{H}}]\nonumber\\&&+(1-\gamma)[|l'_3(f_t)-l'_3(h_t)|\parallel k(x_t^p,\cdot)\parallel_{\mathcal{H}}\nonumber\\&&+|l'_4(f_t)-l'_4(h_t)|\parallel k(x_t^u,\cdot)\parallel_{\mathcal{H}}\nonumber\\&&+|l'_5(f_t)-l'_5(h_t)|\parallel k(x_t^u,\cdot)\parallel_{\mathcal{H}}\nonumber\\&&+|l'_6(f_t)-l'_6(h_t)|\parallel k(x_t^n,\cdot)\parallel_{\mathcal{H}}]   ] \nonumber
	\end{eqnarray}
	\begin{eqnarray}
	&\leq& \kappa^{1/2}\mathbb{E}_{x_{t}^p,x_t^u,x_{t}^u,\omega_t} [\parallel h_t - f^{*} \parallel_{\mathcal{H}} \nonumber \\&&\cdot [\gamma(|l'_1(f_t)-l'_1(h_t)|\nonumber \\&&+|l'_2(f_t)-l'_2(h_t)|)\nonumber\\&&+(1-\gamma)(|l'_3(f_t)-l'_3(h_t)|\nonumber \\&&+|l'_4(f_t)-l'_4(h_t)|\nonumber\\&&+|l'_5(f_t)-l'_5(h_t)|\nonumber \\&&+|l'_6(f_t)-l'_6(h_t)|)]]\nonumber\\ 
	&\leq& \kappa^{1/2} \mathbb{E}_{x_{t}^p,x_t^n,x_{t}^u,\omega_t}[ \| h_t - f^{*} \|_{\mathcal{H}}\nonumber \\&&\cdot[\gamma(L_1|f_t(x_t^p)-h_t(x_t^p)|\nonumber\\&&+L_2|f_t(x_t^n)-h_t(x_t^n)|)\nonumber \\&&+(1-\gamma)(L_3|f_t(x_t^p)-h_t(x_t^p)|\nonumber\\&&+L_4|f_t(x_t^u)-h_t(x_t^u)|\nonumber \\&&+L_5|f_t(x_t^u)-h_t(x_t^u)|\nonumber\\&&+L_6|f_t(x_t^n)-h_t(x_t^n)|)] ] \nonumber \\
	&\leq& \kappa^{1/2}\sqrt{\mathbb{E}_{x_{t}^p,x_t^n,x_{t}^u,\omega_t}[A_t]} \nonumber\\&& \cdot \sqrt{[\gamma(L_1+L_2)+(1-\gamma)(L_3+L_4+L_5+L_6)]^2B_{1,t}^2}\nonumber\\
	&\leq& \kappa^{1/2}LB_{1,t}\sqrt{\mathbb{E}_{x_{t-1}^p,x_{t-1}^n,x_{t-1}^u,\omega_{t-1}}[A_t]} \nonumber
	\end{eqnarray}	
\end{proof}
The first and third inequalities are due to Cauchy-Schwarz Inequality and the second inequality is due to the Assumption 3. And the last step is due to the Lemma \ref{lemma:error_due_random_feature} and the definition of $A_t$.

After proving the Eq. (\ref{Eq_M})-(\ref{Eq_R}) separately, let us denote $e_t= \mathbb{E}_{x_{t}^p,x_{t}^u,\omega_{t}}[A_t]$, given the above bounds, we arrive the following recursion,
\begin{align}
e_{t+1} \leq (1-2\eta_t\lambda)e_t + \kappa M^2 \eta_t^2(1+\lambda c_t)^2+ 2\kappa^{1/2}L\eta_tB_{1,t}\sqrt{e_t}\nonumber
\end{align}

When $\eta_t = \dfrac{\theta}{t}$ with $\theta$ such that $\theta\lambda \in (1,2) \cup \mathbb{Z}_{+}$, from Lemma \ref{lemma:upper_bound}, we have $|a_t^i| \leq \dfrac{\theta}{t}$, $\forall 1 \leq i \leq t$. Consequently, $c_t \leq \theta$ and $B_{1,t}^2 \leq M^2(\kappa+\phi)^2\dfrac{\theta^2}{t-1}$. Applying these bounds to the above recursion, we have
$e_{t+1} \leq (1-\dfrac{2\theta\lambda}{t})e_t + \kappa^2\dfrac{\theta^2}{t^2}(1+\lambda\theta)^2 + 2\kappa^{1/2}\dfrac{\theta}{t}L\eta_t\sqrt{M^2(\kappa+\phi)^2 \dfrac{\theta^2}{t-1}}\sqrt{e_t}$ 
and then it can be rewritten as:
\begin{align}
e_{t+1} \leq (1-\dfrac{2\theta\lambda}{t})e_t+\dfrac{\beta_1}{t}\sqrt{\dfrac{e_t}{t}}+\dfrac{\beta_2}{t^2} \nonumber
\end{align}
where $\beta_1 = 2\sqrt{2}\kappa^{1/2}LM(\kappa+\phi)\theta^2$ ($\sqrt{\dfrac{t}{t-q}}\leq \sqrt{2}$) and $\beta_2 = \kappa M^2\theta^2(1+\lambda\theta)^2$. Reusing Lemma 14 in [Dai \textit{et al.}, 2014] with $\eta = 2\theta\lambda > 1$ leads to
\begin{align}
e_{t+1} \leq \dfrac{Q_1^2}{t} \nonumber
\end{align}
where
\begin{align}
Q_1 = \max\left\{\parallel f^{*} \parallel_{\mathcal{H}},\dfrac{Q_0 + \sqrt{Q_0^2+(2\theta\lambda-1)(1+\theta\lambda)^2\theta^2\kappa M^2}}{2\theta\lambda-1} \right\} \nonumber
\end{align}
and $Q_0 = 2\sqrt{2}\kappa^{1/2}(\kappa+\phi)LM\theta^2$.

\subsection{Proof of Theorem \ref{theorem:Convergence_in_expectation}}
Finally, we give the proof of Theorem \ref{theorem:Convergence_in_expectation}.

\begin{proof}
	Substitute Lemma \ref{lemma:error_due_random_feature} and \ref{lemma:error_due_to_random_data} into Eq. (17), we have that
	\begin{eqnarray}
	\mathbb{E}_{x_{t}^p,x_{t}^u,\omega_t}[|f_t(x)-f^{*}(x)|^2]&\leq&2\mathbb{E}_{x_{t}^p,x_{t}^u,\omega_t}[|f_t(x)-h_{t+1}(x)|^2] \nonumber\\&& + 2\kappa\mathbb{E}_{x_{t}^p,x_{t}^u,\omega_t}[\parallel h_t - f^{*} \parallel_{\mathcal{H}}] \nonumber\\
	&\leq& 2B_{1,t+1}^2+ 2\kappa\dfrac{Q_1^2}{t} \nonumber \\
	&\leq& \dfrac{2C^2+2\kappa Q_1^2}{t} \nonumber
	\end{eqnarray}
	where the last inequality is due to Lemma \ref{lemma:upper_bound}. In this way, we obtain the final result on convergence in expectation. This completes the proof.
\end{proof}

\end{document}